\documentclass[runningheads]{llncs}

\usepackage{eccv}

\usepackage{eccvabbrv}

\usepackage{graphicx}
\usepackage{booktabs}
\usepackage{multirow}
\usepackage{comment}

\usepackage[pagebackref,breaklinks,colorlinks,citecolor=eccvblue]{hyperref}

\usepackage{orcidlink}

\usepackage{enumitem}
\DeclareRobustCommand{\OURS}{Seen2Scene}
\DeclareRobustCommand{\myparagraph}[1]{\vspace{4pt}\noindent\textbf{#1}}

\usepackage[utf8]{inputenc}
\usepackage[most]{tcolorbox}
\tcbuselibrary{skins}
\usepackage{enumitem}
\newtcolorbox{promptbox}[1]{
    colback=blue!5!white,
    colframe=blue!75!black,
    fonttitle=\bfseries,
    coltitle=white,
    title={#1},
    enhanced,
    attach boxed title to top left={yshift=-2mm, xshift=2mm},
    boxed title style={sharp corners=south, colback=blue!90!black},
    rounded corners,
    arc=2mm,
    boxrule=0.5pt,
    left=1mm, right=1mm, top=2mm, bottom=1mm
}

\begin{document}

\title{\OURS: Completing Realistic 3D Scenes with Visibility-Guided Flow}

\titlerunning{\OURS}

\author{Quan Meng$^1$ \and
Yujin Chen$^1$ \and
Lei Li$^2$ \and
Matthias Nie{\ss}ner$^1$ \and
Angela Dai$^1$}

\authorrunning{Q.~Meng et al.}

\institute{$^1$Technical University of Munich \hspace{0.5cm} $^2$University of Virginia\\[2pt]
\url{https://quan-meng.github.io/projects/seen2scene/}}

\maketitle

\begin{figure}[h]
  \centering
  \includegraphics[width=\textwidth]{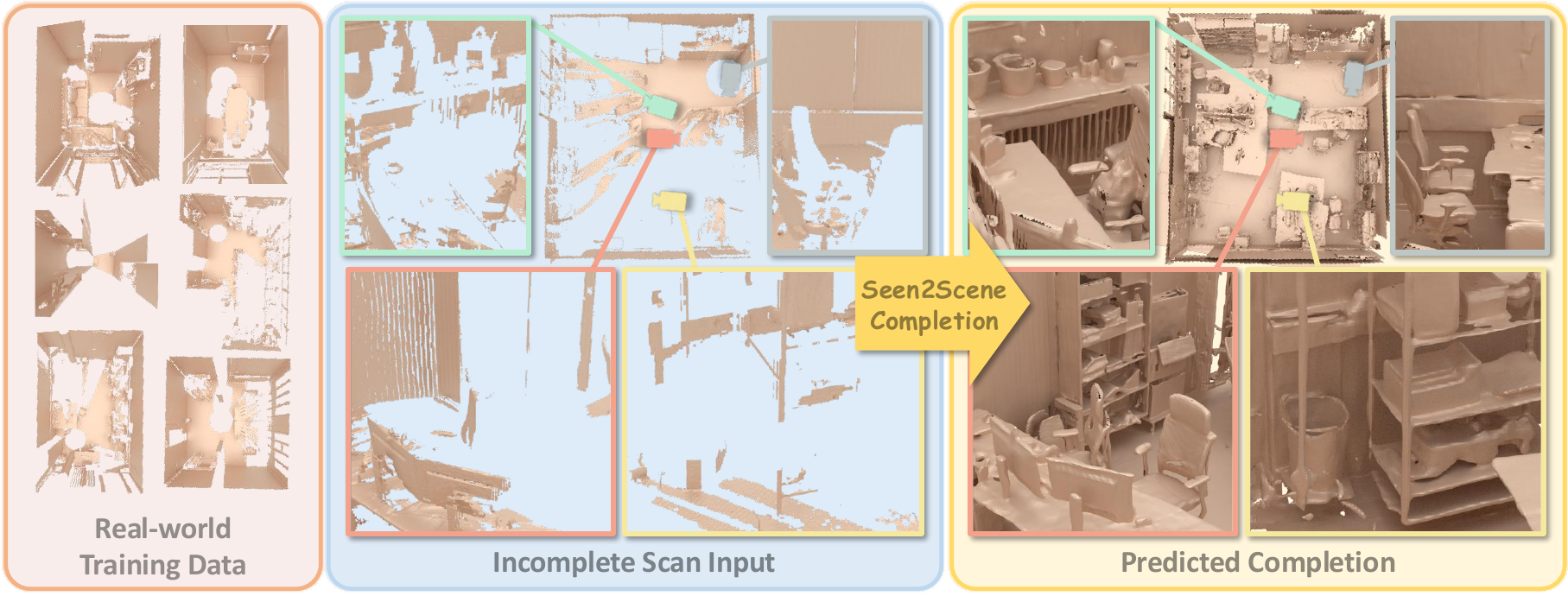}
  \caption{
  \OURS{} is the first visibility-guided flow matching approach for 3D scene completion and generation, trained directly on incomplete, real-world 3D scan data. Our approach predicts high-fidelity, geometrically complete, and structurally coherent scene geometry in unobserved regions, conditioned on observed areas and 3D box-based layouts.
  \vspace{-1.0cm}
  }
  \label{fig:teaser}
\end{figure}

\begin{abstract}
We present \OURS{}, the first flow matching-based approach that trains directly on incomplete, real-world 3D scans for scene completion and generation.
Unlike prior methods that rely on complete and hence synthetic 3D data, our approach introduces visibility-guided flow matching, which explicitly masks out unknown regions in real scans, enabling effective learning from real-world, partial observations.
We represent 3D scenes using truncated signed distance field (TSDF) volumes encoded in sparse grids and employ a sparse transformer to efficiently model complex scene structures while masking unknown regions.
We employ 3D layout boxes as an input conditioning signal, and our approach is flexibly adapted to various other inputs such as text or partial scans.
By learning directly from real-world, incomplete 3D scans, \OURS{} enables realistic 3D scene completion for complex, cluttered real environments.
Experiments demonstrate that our model produces coherent, complete, and realistic 3D scenes, outperforming baselines in completion accuracy and generation quality.
\end{abstract}

\section{Introduction}
\label{sec:intro}

Modern 3D content creation is rapidly shifting from synthetic, isolated objects toward realistic, immersive, and interactive world models.
A significant challenge in this transition is generating complete 3D scenes from partial observations -- a capability increasingly central to content creation, simulation, and planning.
Addressing this challenge calls for 3D generative models that can imagine missing structure while respecting observed geometry and scaling to complex, multi-object indoor scenes.

Recent progress has been made in 3D scene synthesis leveraging the power of deep generative models, in particular, through diffusion-based \cite{wu2024blockfusion, ren2024xcube, meng2025lt3sd, bokhovkin2025scenefactor} and flow matching approaches~\cite{wu2025direct3d, xiang2025structured, xiang2025native}.
However,
training these models heavily relies on synthetic 3D datasets~\cite{fu2021_3dfront, fu2021_3dfuture} that provide complete ground-truth scene geometry for supervision.
This fundamentally limits their ability to generalize to real-world settings.
Unlike synthetic data, real-world scans~\cite{dai2017scannet, yeshwanthliu2023scannetpp, arkitscenes} are inherently incomplete and noisy due to occlusions and limited sensor coverage.
Methods trained exclusively on complete and synthetic 3D scenes often fail to capture the variability, complexity, incompleteness, and off-axis alignments inherent in real scans.

To address these challenges, we propose \OURS{},
a masked flow matching–based approach capable of training directly on incomplete, real-world scans for 3D scene completion and generation.
Our method operates on partial 3D scenes represented as TSDF voxel grids, which naturally encode surfaces while handling unobserved space.
We first train a masked sparse variational autoencoder (VAE) to compress observed TSDF voxel grids into compact geometry latents, masking out unknown regions during training to learn clean latent representations of the observed scene geometry.
A sparse transformer then models the distribution of partial geometry latents using visibility-guided flow matching~\cite{lipman2022flow, liu2022flow}, synthesizing coherent 3D structure conditioned on input 3D box-based layouts.
While designed for partial scan completion, \OURS{} naturally extends to other conditions, such as text prompts, to generate high-fidelity 3D scenes from scratch.

We evaluate our method on the large-scale
ScanNet++~\cite{yeshwanthliu2023scannetpp},
ARKitScenes~\cite{arkitscenes, lazarow2024cubify},
and 3D-FRONT~\cite{fu2021_3dfront} datasets,
where we reconstruct complete indoor scenes from partial TSDF scans and generate 3D scenes from text prompts or layouts.
Our experiments demonstrate that \OURS{} produces complete and realistic 3D scene geometry, outperforming baselines in both geometric completeness and surface fidelity.
These results establish visibility-guided flow matching as an effective paradigm for 3D scene completion from real-world partial data.

In summary, we present the following contributions:
\begin{itemize}[leftmargin=*, labelwidth=1.5em, labelsep=0.5em, topsep=0em, itemsep=0em, parsep=0em]
    \item We present the first masked flow-matching method that can be trained directly on complex, incomplete real-world 3D scans and learn their underlying geometry distributions for accurate scene completion.
    \item We introduce a masked sparse VAE compression that faithfully encodes partial TSDF scans into clean and compact geometry latents, enabling our masked visibility-guided generative modeling to focus on observed scene geometry while effectively handling unknown regions.
    \item Our method can be easily adapted to generate complete, high-fidelity 3D scenes from scratch conditioned on text prompts and/or object layouts, demonstrating strong flexibility and generalization beyond scene completion.

\end{itemize}

\section{Related Work}
\label{sec:related_works}

\myparagraph{Score-Based 3D Shape Generation.}
Following the success of diffusion~\cite{ho2020denoising,rombach2021highresolution} and flow matching~\cite{lipman2022flow, liu2022flow} in the image domain, 3D shape generative methods have adopted this generative paradigm for various 3D shape representations: point clouds~\cite{zhou20213d}, learned latent codes~\cite{vahdat2022lion}, and implicit function parameters~\cite{chou2023diffusion,jun2023shap,erkocc2023hyperdiffusion,zhang2025dnf}.
More recently, structured sparse latents~\cite{xiang2025structured,xiang2025native, li2025sparc3d, ren2024xcube} and scalable 3D diffusion transformers~\cite{chen20253dtopia, wu2025direct3d} have pushed both shape quality and resolution further.
However, these methods focus on object-centric generation, relying on datasets of curated synthetic object assets.
Our method builds upon flow matching and scales 3D generative models to unstructured, incomplete, large-scale real-world data such as RGB-D scans.

\myparagraph{3D Scan Completion.}
3D scan completion aims to reconstruct the full geometry representing a scene from partial observations such as RGB-D inputs or sparse LiDAR scans. Early approaches employed volumetric 3D CNNs \cite{dai2018scancomplete,dai2020sg,huang2023nksr}, with SG-NN~\cite{dai2020sg} introducing self-supervised sparse generative modeling from partial scans. Subsequent works improved structural priors and robustness via sketch-aware supervision \cite{chen20203d}, anisotropic convolutions \cite{li2020anisotropic}, and uncertainty modeling \cite{cao2024pasco}. To mitigate occlusion, camera-based methods exploit temporal context \cite{li2024hierarchical}, virtual multi-view augmentation \cite{selvakumar2025fake, huang2024ssr}, and pseudo-future modeling \cite{lu2025one}. \cite{li2024sscbench, han2025voic, xi2026flowssc} infer a complete 3D volumetric representation of both semantics and geometry from a single RGB image.
However, these works view completion from a regression standpoint. Generative completion works~\cite{chu2023diffcomplete,schaefer2024sc,zhang2024outdoor} incorporate strong generation ability for completion task. We treat scene completion as sparse-scan-conditioned generation, which enables more direct fine-tuning of pretrained generative foundation models, achieving significant generalization to unobserved regions while keeping fidelity to observations.

\myparagraph{3D Scene Generation.}
3D scene generation aims to capture global layout structure and local geometry~\cite{infinigen2023infinite, infinigen2024indoors}.
Layout-conditioned approaches~\cite{tang2024diffuscene, zhai2024echoscene, sun2025semlayoutdiff, chen2025layout2scene, fang2025spatialgen} learn to synthesize scenes given spatial arrangements of objects.
To achieve scalability in expansive environments, recent methods focus on chunk-based generation and outpainting~\cite{ren2024xcube, ju2024diffindscene, wu2024blockfusion, meng2025lt3sd, bokhovkin2025scenefactor, hollein2023text2room}.
While effective in high-resolution synthesis, these models are trained using complete, synthetic 3D scene datasets limited in diversity and asset categories, limiting their ability to bridge the domain gap to real-world 3D scan data.
Recently, agentic frameworks~\cite{yang2025sceneweaver, pfaff2026scenesmith, wu2025diorama, ling2025scenethesis} have introduced LLM/VLM planners to produce object-based indoor 3D scenes. Although these methods excel in high-level planning, they often rely on asset retrieval or simplified layout assumptions (\eg, axis-aligned), struggling to synthesize complex, high-fidelity geometry reflective of real-world, cluttered environments~\cite{dai2017scannet,arkitscenes,yeshwanthliu2023scannetpp}.
In contrast, we introduce visibility-aware masked flow matching that learns directly from incomplete, noisy real scans.

\section{Method}
\label{sec:methods}
\begin{figure}[t]
  \centering
  \includegraphics[width=1.0\linewidth]{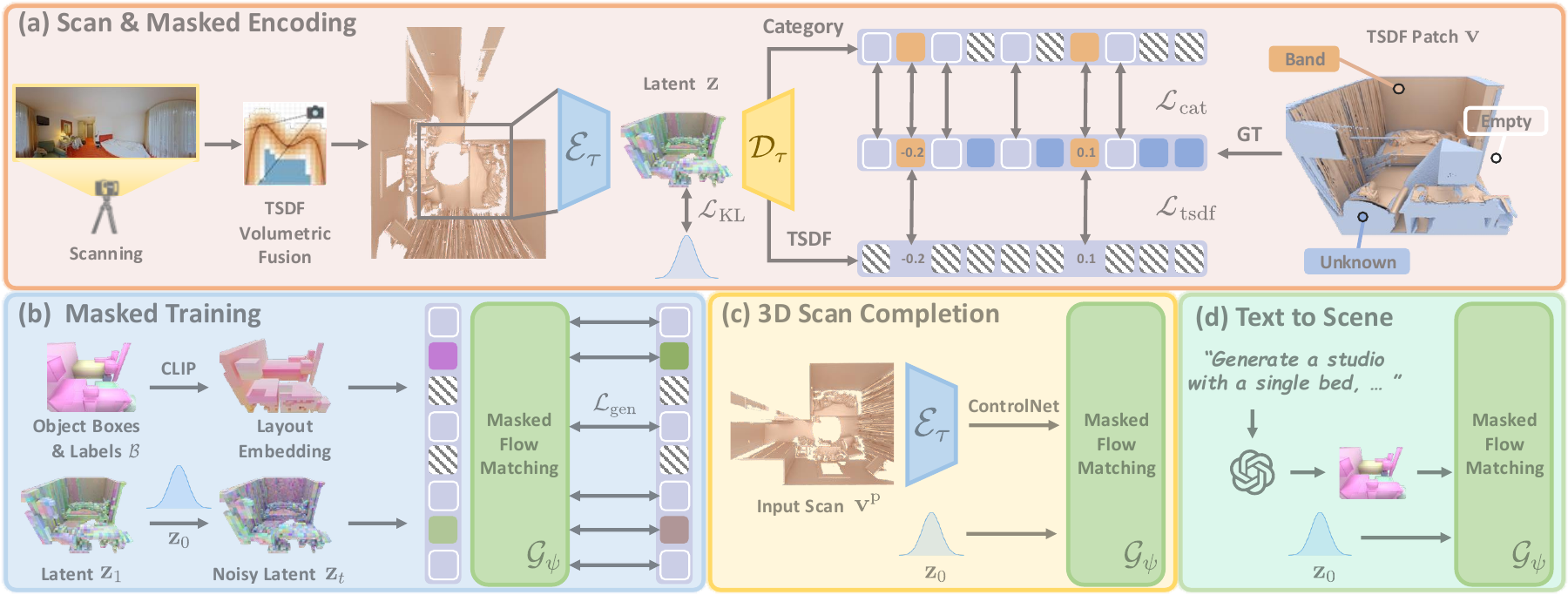}
  \caption{\textbf{Overview of \OURS{}.}
  We introduce visibility-guided flow matching for
  modeling the distribution of
  TSDF partial scans.
  \textbf{(a)} Partial scan TSDF patches $\mathbf{v}$ are encoded by a masked sparse VAE ($\mathcal{E}_{\tau}, \mathcal{D}_{\tau}$) into latent representations $\mathbf{z}$, masking out unknown regions unseen by the camera.
  \textbf{(b)} A sparse transformer $\mathcal{G}_\psi$ conditioned on 3D layout boxes $\mathcal{B}$ is trained with masked flow matching on surface and empty region tokens.
  \textbf{(c)} We fine-tune $\mathcal{G}_\psi$  for scan completion by injecting partial scan inputs $\mathbf{v}^{\mathrm{p}}$ via ControlNet~\cite{zhang2023adding}. \textbf{(d)} $\mathcal{G}_\psi$ can also be flexibly adapted for text or layout-conditioned 3D scene generation from scratch.
  }
  \label{fig:pipeline}
  \vspace{-0.5cm}
\end{figure}

We propose a new generative approach, \OURS{}, which synthesizes complete, high-fidelity 3D scenes from partial, real-world scans.
Key to our method is visibility-guided flow matching, which enables generative modeling on incomplete 3D scan data by focusing on observed scene geometry while remaining agnostic to unknown areas.

An overview of our approach is illustrated in \cref{fig:pipeline}. Input 3D scans are represented as TSDF volumes reconstructed via volumetric fusion \cite{curless1996volumetric} (\cref{sec:data_gen}).
We then sample and encode patches, denoted as $\mathbf{v}$, from these partial TSDF into compact geometry latents $\mathbf{z}$ using a masked sparse VAE (\cref{sec:sparse_vae}).
As the TSDF representation inherently encodes which regions are
unknown (\ie, unseen by the scanning camera, with signed distance of $-3 \times \text{voxel size}$), these regions are masked out during their encoding.
A sparse transformer $\mathcal{G}_\psi$ is then trained with visibility-aware masked flow matching to generate geometry latents from Gaussian noise conditioned on 3D bounding box layouts, with supervision only on observed regions, while generated geometry latents for unknown regions are masked and excluded from the flow matching loss (\cref{sec:masked_fm}).
This flow matching model then serves as the backbone for 3D scene completion and generation.
To complete partial scans, we fine-tune the pretrained generative model by injecting input partial scan TSDFs via a ControlNet~\cite{zhang2023adding} branch, which guides generation of unknown regions while preserving observed geometry (\cref{sec:3d_scene_gen}).

\subsection{Visibility-Aware Structured Scene Representation}
\label{sec:data_gen}

To learn directly on partial scans of real-world environments, at each iteration, we randomly sample a patch $\mathbf{v}$ of resolution $256^3$ within the scene. The patch $\mathbf{v}$ is stored as a TSDF grid reconstructed from raw depth sequences of the environment via volumetric fusion~\cite{curless1996volumetric}, as shown in \cref{fig:pipeline}(a).

This volumetric representation naturally handles the incompleteness of real-world scans and encodes visibility information. Voxels with TSDF values greater than the negative truncation bound encode known geometry observed by the camera, while voxels at the sentinel value of $-3\times\text{voxel size}$ indicate unknown regions unseen by the camera.
We explicitly leverage this visibility information to mask out unknown regions during both the scene latent encoding and generative modeling stages. This ensures that the model is supervised only by valid geometry observations, preventing it from learning the ambiguity from unobserved areas to enable more generalized completion.

\subsection{Latent Encoding with Masked Sparse VAE}
\label{sec:sparse_vae}

To efficiently model high-resolution 3D scene geometry, we first compress the TSDF chunks $\mathbf{v}$ into a compact latent representation $\mathbf{z}$. We design a visibility-aware masked sparse VAE, as illustrated in \cref{fig:pipeline}(a), which explicitly handles the inherent sparsity and incompleteness of real-world scans.

This sparse VAE consists of a sparse encoder $\mathcal{E}_\tau$ and a sparse decoder $\mathcal{D}_\tau$, parameterized by $\tau$, both built with residual sparse convolutional blocks \cite{williams2024fvdb}.
Given the TSDF patch $\mathbf{v}$, the encoder $\mathcal{E}_\tau$ progressively downsamples the sparse voxel features and maps them to a latent distribution $q(\mathbf{z} \mid \mathbf{v})$, regularized toward a standard normal prior $p(\mathbf{z})$ via a KL divergence loss $\mathcal{L}_{\mathrm{KL}}$.
The decoder $\mathcal{D}_\tau$ then reconstructs the input partial geometry from sampled latents through consecutive upsampling.
To predict sparse 3D surfaces, the decoder has two prediction heads: a \emph{category head} that classifies each voxel as surface (i.e., voxels with distance smaller than $3 \times \text{voxel size}$, colorized as orange in \cref{fig:pipeline}) or empty (i.e., voxels with signed distance of $3 \times \text{voxel size}$, colorized as white in \cref{fig:pipeline}), and a \emph{TSDF head} that regresses signed-distance values only for surface voxels.

We adopt a masked training strategy~\cite{dai2020sg} to accommodate the incompleteness of real-world scans in unobserved regions (highlighted in blue in \cref{fig:pipeline}). Such incompleteness arises from inevitable occlusion and reflective surfaces — for instance, regions beneath tables and in front of windows are largely unscanned, as illustrated in \cref{fig:pipeline} (top right).
Training losses are computed only over valid (known observed) voxels, encouraging the model to learn a more complete latent representation from partial observations while ignoring unobserved regions. The total loss is
\begin{equation}
\mathcal{L}_{\mathrm{vae}} = \mathcal{L}_{\mathrm{tsdf}} + \mathcal{L}_{\mathrm{cat}} + \lambda_{\mathrm{KL}} \, D_{\mathrm{KL}}\!\bigl(q(\mathbf{z} \mid \mathbf{v})\,\|\,p(\mathbf{z})\bigr),
\end{equation}
where $\mathcal{L}_{\mathrm{tsdf}} = \frac{1}{|m_s|}\sum_{j \in m_s}\lvert \hat{v}_j - v_j \rvert$ is the masked $\ell_1$ reconstruction loss, $\mathcal{L}_{\mathrm{cat}} = -\frac{1}{|m|}\sum_{j \in m} [y_j \log \hat{y}_j + (1-y_j)\log(1-\hat{y}_j)]$ is the masked binary cross-entropy for voxel occupancy classification, where $j$ indexes voxels, $1$ represents surface, and $0$ represents empty space. The mask $m_s$ denotes the set of surface voxels, $m$ denotes all known voxels (surface $m_s$ and empty space $m_e$), the loss weight $\lambda_{\mathrm{KL}}$ weights the KL regularization toward the standard normal prior.

\subsection{Visibility-Aware Masked Flow Matching}
\label{sec:masked_fm}
With the observed scene regions encoded into a compact geometry latent space, we train a conditional generative model $\mathcal{G}_\psi$, parameterized by $\psi$, to generate these geometry latents, conditioned on a 3D semantic scene layout, as shown in \cref{fig:pipeline}(b).
By observing many partial examples of objects (\eg, chairs, cabinets) from different angles across diverse real-world 3D scenes, the model learns the underlying distribution of object and scene geometry.
The 3D semantic layout, specifying object spatial extents and categories, provides crucial coarse structural cues for inferring plausible geometry in unobserved regions.

We represent the 3D semantic layout condition as a set of axis-aligned object bounding boxes with semantic labels $\mathcal{B} = \{\mathbf{b}_k = (\mathbf{c}_k, \mathbf{s}_k, l_k)\}_{k=1}^{K}$, where each box $\mathbf{b}_k$ is defined by its centroid $\mathbf{c}_k$, size $\mathbf{s}_k$, and semantic label $l_k$, and $k$ indexes objects in the scene.
For each scene chunk, we collect all intersecting object bounding boxes to form $\mathcal{B}$.
We encode each instance's semantic label with a CLIP text encoder \cite{radford2021clip}. The resulting semantic features are painted into a 3D layout map according to each box's spatial extent.
This layout map is spatially aligned with the geometry latent grid and embedded into a layout token sequence, serving as the conditioning signal.

We formulate scene geometry generation with flow matching~\cite{lipman2022flow, liu2022flow}, using a sparse transformer as the generative model backbone $\mathcal{G}_\psi$.
The model
learns a velocity field that transports samples from a Gaussian prior $p_0$ to the target geometry distribution $p_{\text{geo}}$. Self-attention~\cite{dao2022flashattention, dao2023flashattention2} is applied across all geometry and layout tokens jointly. The training loss is
\begin{equation}
    \mathcal{L}_{\mathrm{gen}}=\mathbb{E}_{\mathbf{z}_0 \sim p_0,\, \mathbf{z}_1 \sim p_{\text{geo}},\, t \sim \mathcal{U}(0,1)}\left[\left\|\mathcal{G}_\psi\left(\mathbf{z}_t, t, \mathcal{B}\right)-\left(\mathbf{z}_1-\mathbf{z}_0\right)\right\|^2\right],
\end{equation}
where $\mathbf{z}_t = (1-t)\,\mathbf{z}_0 + t\,\mathbf{z}_1$ is the linear interpolant between noise $\mathbf{z}_0$ and target latent $\mathbf{z}_1$, and the model predicts the velocity $\mathbf{z}_1 - \mathbf{z}_0$.
During training, geometry tokens corresponding to unobserved regions (as indicated by the visibility mask from the sparse VAE) are excluded from the flow matching objective, so the model learns to generate geometry only for observed regions while remaining agnostic to missing areas.
Following classifier-free guidance~\cite{ho2022classifier}, we randomly drop $\mathcal{B}$ during training, making the layout condition optional at test time, as validated by our experiments.
We provide more comprehensive training details in the supplementary material.

\subsection{Scan Completion and Multi-modal Generation}
\label{sec:3d_scene_gen}

\myparagraph{3D Scan Completion.}
To infer the missing structure of input 3D scans, we leverage the generative priors learned from partially observed, real-world scene geometry (\cref{sec:masked_fm}), as shown in \cref{fig:pipeline}(c).
We adopt a self-supervised formulation~\cite{dai2020sg} by fine-tuning the pretrained generative model $\mathcal{G}_\psi$ on existing 3D scan training data, but with varying levels of incompleteness.
Concretely, we remove a portion of the depth frames from an existing partial scan $\mathbf{v}$ to create a more incomplete variant $\mathbf{v}^{\mathrm{p}}$, serving as input condition. The original partial scan $\mathbf{v}$ is used as the target for supervision.
This encourages the model to learn to correlated different levels of incompleteness within the same scene and predict the removed geometry.

We add a ControlNet~\cite{zhang2023adding} branch $\mathcal{C}_\phi$, parameterized by $\phi$ which is initialized from the pretrained weights of $\mathcal{G}_\psi$, to inject the more incomplete scan condition $\mathbf{v}^{\mathrm{p}}$ into the flow matching model $\mathcal{G}_\psi$ during fine-tuning, without modifying the pretrained flow matching weights.
This conditioning provides the scene context and geometric cues for the synthesis of unobserved regions, while ensuring observed areas are preserved.

Formally, the more partial scan $\mathbf{v}^{\mathrm{p}}$ is first encoded by the frozen VAE encoder $\mathcal{E}_\tau$ into a latent condition $\mathbf{z}^{\mathrm{p}} = \mathcal{E}_\tau(\mathbf{v}^{\mathrm{p}})$. The ControlNet branch $\mathcal{C}_\phi$ processes $\mathbf{z}^{\mathrm{p}}$ and injects per-layer control signals into the frozen base model~\cite{zhang2023adding}. The fine-tuning objective follows the same flow matching loss:
\begin{equation}
    \mathcal{L}_{\mathrm{sc}} = \mathbb{E}_{\mathbf{z}_0,\, \mathbf{z}_1,\, t}\left[\left\|\mathcal{G}_\psi\!\left(\mathbf{z}_t, t, \mathcal{B}, \mathcal{C}_\phi(\mathbf{z}^{\mathrm{p}})\right) - \left(\mathbf{z}_1 - \mathbf{z}_0\right)\right\|^2\right],
\end{equation}
where only the ControlNet parameters $\phi$ are optimized while the pretrained weights of $\mathcal{G}_\psi$ remain frozen.
During training, the layout condition $\mathcal{B}$ is derived from the ground-truth semantic annotations available in the datasets.
At inference on novel scans, $\mathcal{B}$ is optional as generation in \cref{sec:masked_fm}, as validated by our experiments.

\myparagraph{Text-to-3D Scene Generation.}
While optimized for real-world scene completion, our generative model backbone $\mathcal{G}_\psi$ is flexible and generalizes to 3D layouts converted from other conditioning modalities, such as natural language prompts, enabling high-fidelity scene generation from scratch, as shown in \cref{fig:pipeline}(d).
For instance, a given text prompt like ``\textit{Generate a studio with a single bed, \ldots}'' can be first analyzed by a large language model (LLM)~\cite{achiam2023gpt} and then translated into a structured set of semantic labels and 3D bounding boxes. To improve robustness to diverse object labels, we augment each label with LLM-generated synonyms, abbreviations, plurals, and typos.
Alternatively, users can directly specify or adjust 3D layout information to define scene structure.

\myparagraph{Generating Coherent, Large-Scale 3D Scenes.}
While our approach is trained on chunks for efficiency, we can directly generalize to large-scale scan completion or generation of arbitrary-sized scenes by applying MultiDiffusion~\cite{bar2023multidiffusion}, similar to LT3SD~\cite{meng2025lt3sd}.
Specifically, we employ a tiled generation strategy across multiple overlapping $256^3$ chunks.
At each timestep, we split the 3D scene into chunks with an overlap ratio of $0.2$ with respect to the chunk size and take each denoising step on all chunks simultaneously.
Then, an average-based blending strategy is applied for the overlapping regions to ensure seamless fusion between adjacent chunks.
Finally, the generated chunk-wise latent representations are decoded and fused into a global large-scale and multi-room 3D geometry.

\section{Experiments}
\label{sec:experiments}
\begin{figure}[t!]
  \centering
  \includegraphics[width=1.0\linewidth]{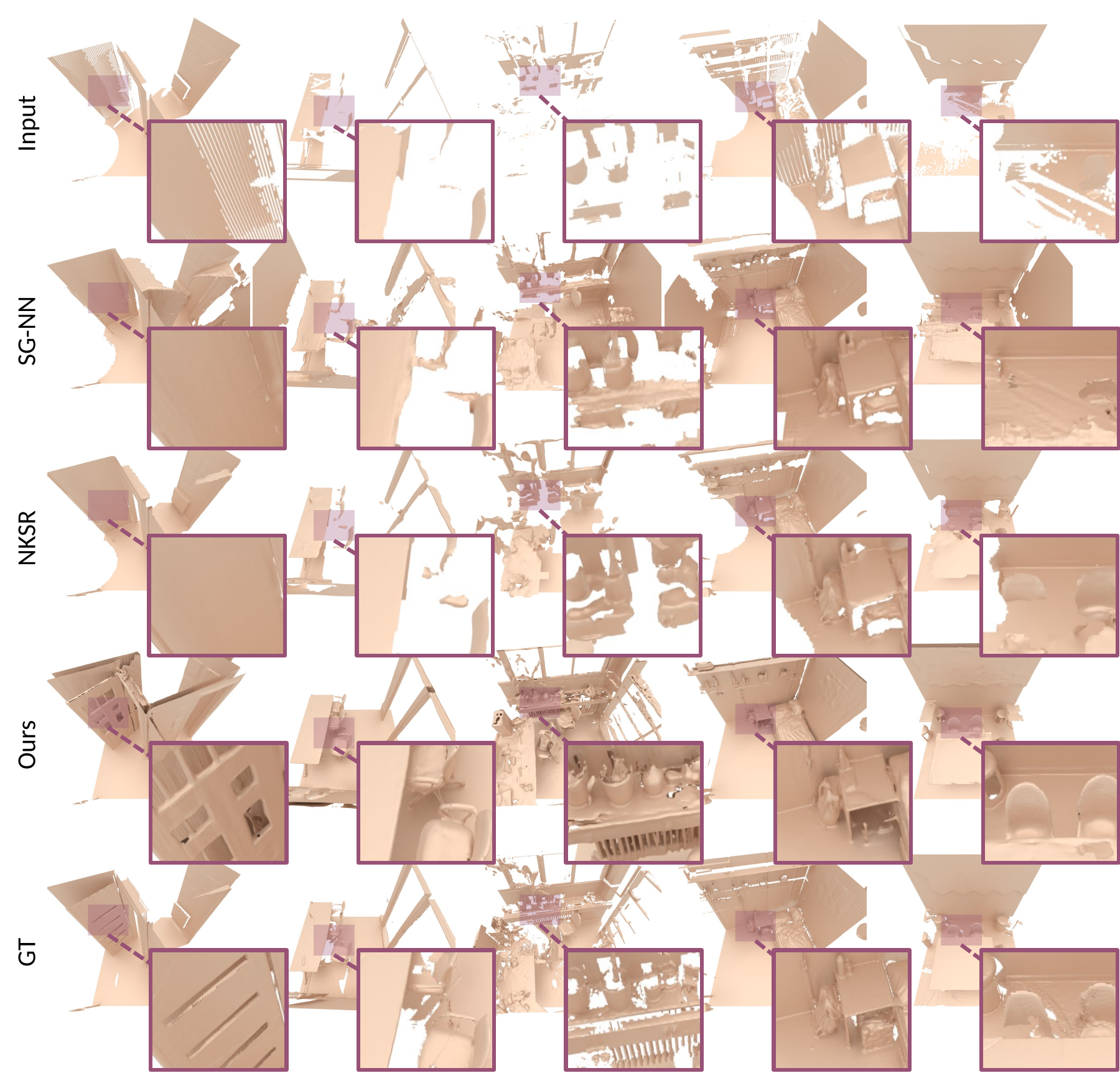}
  \caption{\textbf{Qualitative comparison on 3D scan completion.} Given real-world partial depth scans from ScanNet++~\cite{yeshwanthliu2023scannetpp} and ARKitScenes~\cite{arkitscenes}, \OURS{} learns realistic, high-fidelity priors from incomplete real-scan data to produce much more complete, detailed geometry than baselines.
  }
  \label{fig:sc_baselines}
  \vspace{-0.5cm}
\end{figure}

\subsection{Implementation Details}
\label{sec:impl_details}

We train on ScanNet++~\cite{yeshwanthliu2023scannetpp}, ARKitScenes~\cite{arkitscenes,lazarow2024cubify}, and 3D-FRONT~\cite{fu2021_3dfront}, which together provide large-scale real-world and synthetic indoor scenes with object annotations. We apply masked training on all three datasets. For 3D-FRONT, we generate synthetic partial scan data by randomly sampling cameras~\cite{Denninger2023} and render depth maps to simulate real-world scanning. TSDF volumes~\cite{curless1996volumetric} are reconstructed using VDBFusion~\cite{vizzo2022sensors} with a voxel size of $1.1\,\text{cm}$ and truncation of $3\times$ voxel size,
then divided into chunks of resolution $256^3$.
Object names are encoded with CLIP ViT-B/32~\cite{radford2021clip}.
The sparse VAE uses fVDB~\cite{williams2024fvdb} residual convolutional blocks
with $8\times$ downsampling, and $8$ latent channels.
The SparseDiT has $28$ transformer blocks and uses RoPE~\cite{su2024rope} positional encoding.
Both stages are trained with AdamW \cite{loshchilov2017decoupled}, a learning rate of $10^{-4}$, cosine annealing with $1000$ warmup steps,
and a batch size of $64$ on 4 NVIDIA H100 GPUs with BF16 mixed precision.
At inference, we use $50$-step Euler sampling with classifier-free guidance~\cite{ho2022classifier} at scale $3.0$.
More data processing details are in the supplementary.

\begin{table}[t]
\centering
\caption{\textbf{Quantitative evaluation of scan completion.} We compare \OURS{} with SG-NN~\cite{dai2020sg} and NKSR~\cite{huang2023nksr}: CD on 3D-FRONT~\cite{fu2021_3dfront}, and L2, TMD, and U3D-FPD on ScanNet++~\cite{yeshwanthliu2023scannetpp}, ARKitScenes~\cite{arkitscenes}, and 3D-FRONT~\cite{fu2021_3dfront}.}
\label{tab:scene_completion}
\resizebox{\linewidth}{!}{%
\small
\setlength{\tabcolsep}{5pt}
\begin{tabular}{l c c c c}
\toprule
\textbf{Method} & \textbf{CD}{\scriptsize$\times10^{-2}$}$\downarrow$ & \textbf{L2}{\scriptsize$\times10^{-4}$}$\downarrow$ & \textbf{TMD}{\scriptsize$\times10^{-2}$}$\uparrow$ & \textbf{U3D-FPD}{\scriptsize$\times10^{-2}$}$\downarrow$ \\
\midrule
SG-NN~\cite{dai2020sg}  & 10.77 & 1.90 & 0.00   & 15.41   \\
NKSR~\cite{huang2023nksr}  & 5.22  &  2.55  & 0.00  & 21.78   \\
\OURS~{\scriptsize(w/o bbox)}        & \textbf{1.92}  & 0.53 & 1.69   & 9.57   \\
\OURS         & 2.05   & \textbf{0.51} & \textbf{3.33} &  \textbf{7.79} \\
\bottomrule
\end{tabular}%
}
\vspace{-0.5cm}
\end{table}

\begin{figure}[h!]
  \centering
  \includegraphics[width=1.0\linewidth]{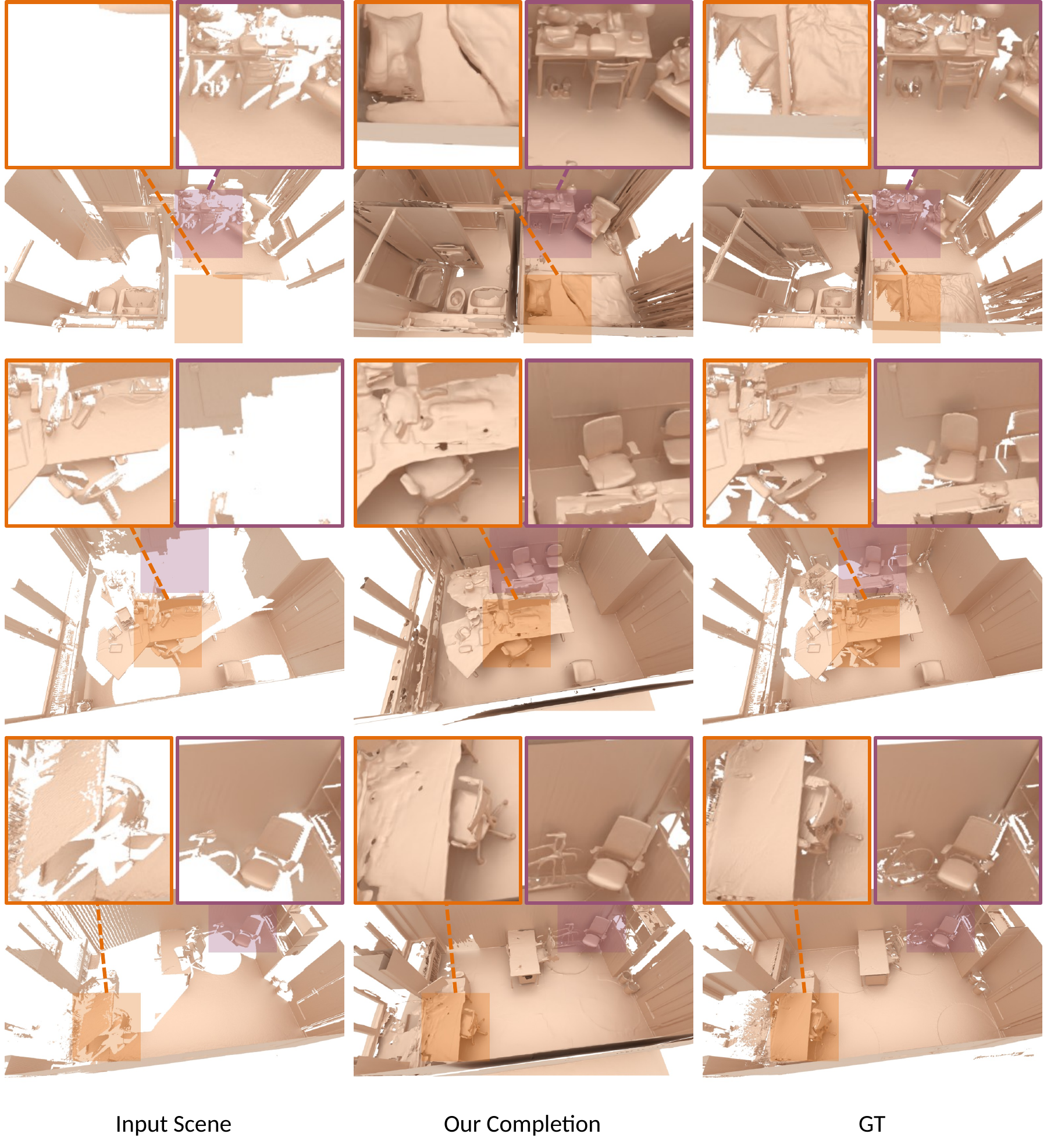}
  \vspace{-0.6cm}
  \caption{
  \textbf{3D scene completion.} \OURS{} can complete large-scale scenes by generating geometry for unobserved regions across multiple chunks, guided by partial scan conditions injected into the ControlNet branch. Results are shown on ScanNet++~\cite{yeshwanthliu2023scannetpp} and ARKitScenes~\cite{arkitscenes}.
  }
  \label{fig:exp_large_sc}
  \vspace{-0.6cm}
\end{figure}

\myparagraph{Evaluation Metrics.}
For scan completion, we compare with the overall $4,000$ samples from the three datasets. We compute Chamfer distance (CD) between uniformly sampled pointclouds from the meshes, and metrics that can be computed only in the observed region of target scans: L2 distance and Total Mutual Difference (TMD) which computes diversity across two completions via pairwise CD. Note that the CD metric is only computed on 3D-FRONT~\cite{fu2021_3dfront} which contains complete mesh ground truth, while other metrics are computed on ScanNet++~\cite{yeshwanthliu2023scannetpp}, ARKitScenes~\cite{arkitscenes,lazarow2024cubify}, and 3D-FRONT~\cite{fu2021_3dfront}.
For scene generation, we report DINOv2-FID~\cite{oquab2023dinov2} computed from $4,000$ patches with 30 views per patch with different elevations, Uni3D Fr\'echet Point Cloud Distance (U3D-FPD)~\cite{zhouuni3d}, which measures distributional similarity in a semantically aligned 3D feature space, and a vision-language model (VLM)-based perceptual score that aggregates geometric quality, structural connectivity, semantic coherence, and content diversity using Qwen3-VL-8B-Instruct~\cite{yang2025qwen3} (prompt details in the supplementary). Note that all the baselines are developed for 3D-FRONT only, so we compute all the metrics only on 3D-FRONT.

\subsection{3D Scan Completion}

Given partial depth scans, \OURS{} completes the scene by generating plausible geometry for unobserved regions via the ControlNet branch. We compare SG-NN~\cite{dai2020sg}, NKSR~\cite{huang2023nksr}, ours without object bounding boxes $\mathcal{B}$, and ours with $\mathcal{B}$ in \cref{tab:scene_completion}.
Our model without bounding boxes achieves comparable performance to its bounding box-conditioned counterpart, and even yields better overall structural correctness (CD). This may be because bounding box information is redundant and potentially disruptive for completing small missing regions (e.g., holes on object surfaces), yet beneficial for recovering large unobserved regions — as illustrated by the two office chairs at the top of the second sample in~\cref{fig:exp_large_sc}.
Since SG-NN and NKSR are both deterministic, their TMD values are zero.
\cref{fig:sc_baselines} shows qualitative comparisons: our method produces more complete and structurally coherent completions.
\OURS{} can also complete large-scale scenes by generating geometry for unobserved regions across multiple chunks. \cref{fig:exp_large_sc} shows that our method generates high-fidelity, geometrically complete, and globally consistent reconstructions in multi-room environments. We provide more results in the supplementary.

\subsection{3D Scene Generation}
\label{sec:scene_gen}
We evaluate layout-conditioned 3D scene generation against BlockFusion~\cite{wu2024blockfusion}, LT3SD~\cite{meng2025lt3sd}, and WorldGrow~\cite{worldgrow2025}.
Since LT3SD and WorldGrow are unconditional generators while BlockFusion and ours are layout-conditioned, we ensure a fair comparison by splitting the 3D-FRONT test set into two disjoint subsets: layouts from the first subset serve as conditions for BlockFusion and \OURS{}, while the second subset is used as the reference distribution for all methods.

\begin{figure}[t]
  \centering
  \includegraphics[width=1.0\linewidth]{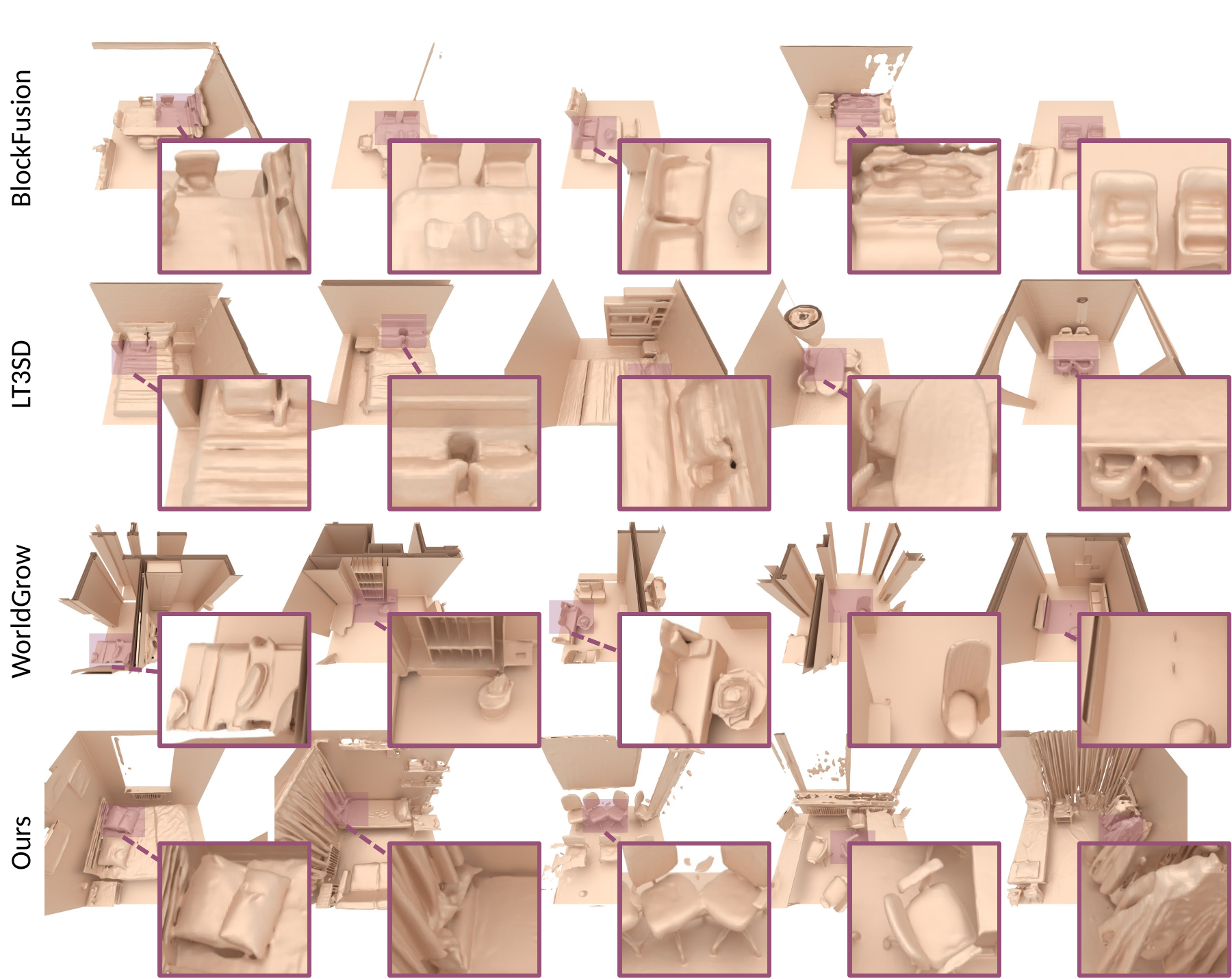}
  \caption{\textbf{Qualitative comparison on layout-conditioned 3D scene generation.} Our method produces more geometrically detailed and semantically coherent scenes compared to BlockFusion~\cite{wu2024blockfusion}, LT3SD~\cite{meng2025lt3sd}, and WorldGrow~\cite{worldgrow2025}. Layouts are from ScanNet++~\cite{yeshwanthliu2023scannetpp} and ARKitScenes~\cite{arkitscenes}.}
  \label{fig:sg_baselines}
\vspace{-0.5cm}
\end{figure}

\begin{table}[h!]
\centering
\caption{\textbf{Quantitative comparison on 3D scene generation.} We report DINOv2 Fr\'echet Inception Distance (FID), Uni3D Fr\'echet Point Cloud Distance (FPD), and VLM Quality Score assessed by Qwen3-VL on a 0--10 scale 3D-FRONT~\cite{fu2021_3dfront} only. }
\label{tab:baselines}
\small
\begin{tabular}{l @{\hspace{10pt}} c @{\hspace{10pt}} c @{\hspace{10pt}} c}
\toprule
\textbf{Method} & \textbf{DINOv2-FID}{\scriptsize$\times10^{2}$}$\downarrow$ & \textbf{U3D-FPD}{\scriptsize$\times10^{2}$}$\downarrow$ & \textbf{VLM Score}$\uparrow$ \\
\midrule
BlockFusion~\cite{wu2024blockfusion}  & 12.92  & 28.30  & 3.51 \\
LT3SD~\cite{meng2025lt3sd}            & 3.51  & 28.55  & 5.20 \\
WorldGrow~\cite{worldgrow2025}        & 4.44  & 50.83  & 4.69 \\
\midrule
\OURS~{\scriptsize(Synthetic only)}   & \textbf{2.73}  & \textbf{11.55} & 5.61 \\
\bottomrule
\end{tabular}
\end{table}

As shown in \cref{tab:baselines}, \OURS{} achieves the best performance across all three metrics.
Our method obtains the lowest DINOv2-FID, indicating that our generated scenes are perceptually closest to high-quality ground-truth meshes.
The U3D-FPD gap demonstrates that the generated 3D geometry is distributionally well-aligned with ground-truth scenes in a semantically meaningful feature space.
The combination of our sparse transformer operating on compact TSDF latents and CLIP-based layout conditioning jointly captures both fine-grained geometric structure and high-level semantic consistency.
This improvement can be attributed to our visibility-guided flow matching, which learns geometry distributions directly from real-world partial scans rather than relying solely on synthetic data, producing more realistic surface details and object arrangements.
Qualitatively, \cref{fig:sg_baselines} shows that our method generates more geometrically detailed, semantically coherent, and realistic scenes, whereas baselines tend to produce over-smoothed surfaces, noticeable artifacts, axis-aligned objects, and clean layout.

\begin{figure}[t]
  \centering
  \includegraphics[width=1.0\linewidth]{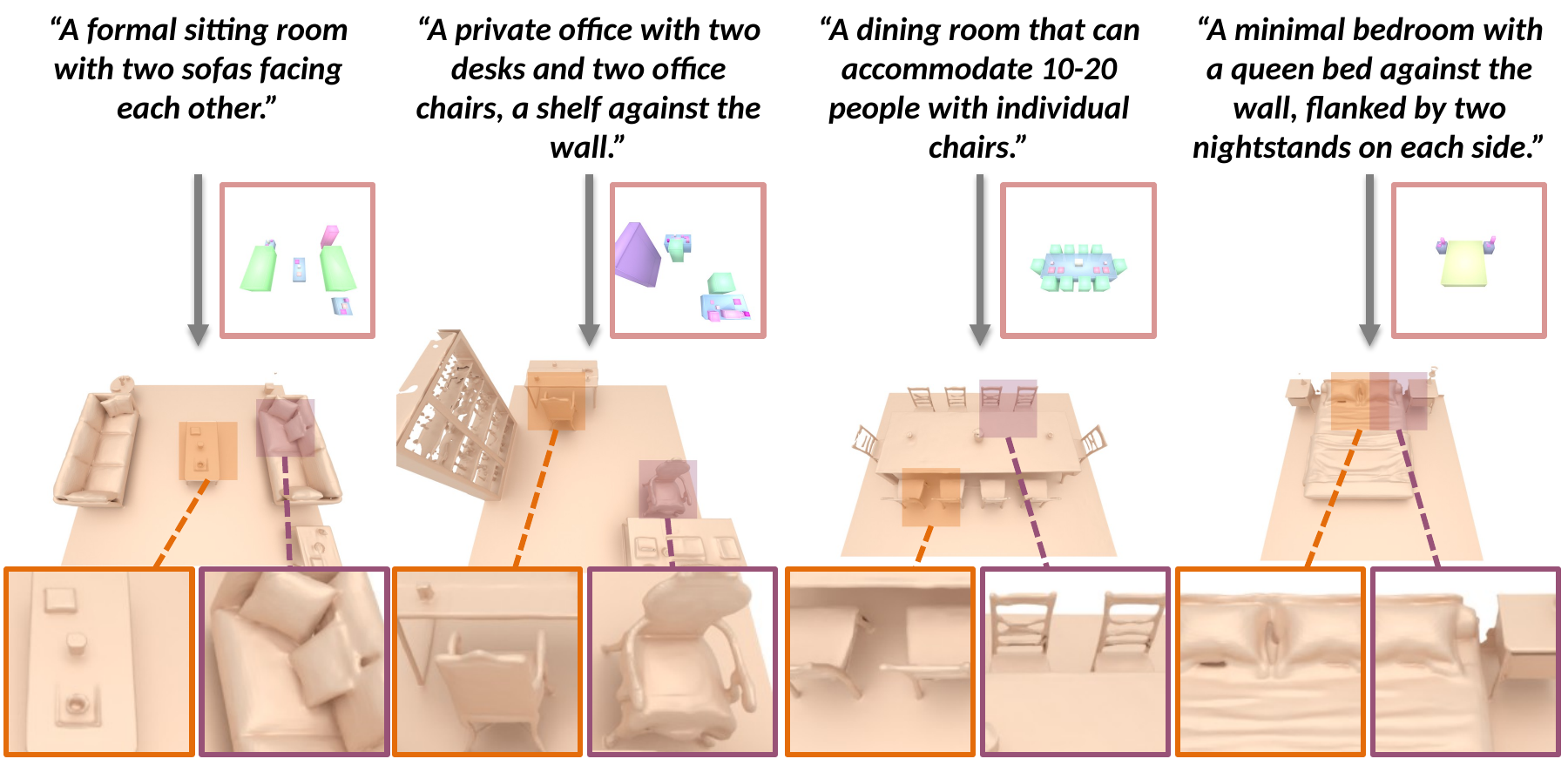}
  \caption{\textbf{Text to scene.} Given a text description, \OURS{} generates a 3D object layout first via an LLM and then synthesizes high-fidelity scene geometry from scratch conditioned on the layout.
  }
  \label{fig:exp_text2scene}
  \vspace{-0.5cm}
\end{figure}

\OURS{} generates 3D scenes from scratch conditioned on text descriptions by first producing an object layout (3D bounding boxes with semantic labels) and then synthesizing geometry conditioned on this layout (\cref{sec:3d_scene_gen}).
\cref{fig:exp_text2scene} shows that our approach robustly generalizes to 3D layouts inferred by an LLM.

\subsection{Ablation Studies}
\label{sec:ablation}
\begin{figure}[t]
  \centering
  \includegraphics[width=1.0\linewidth]{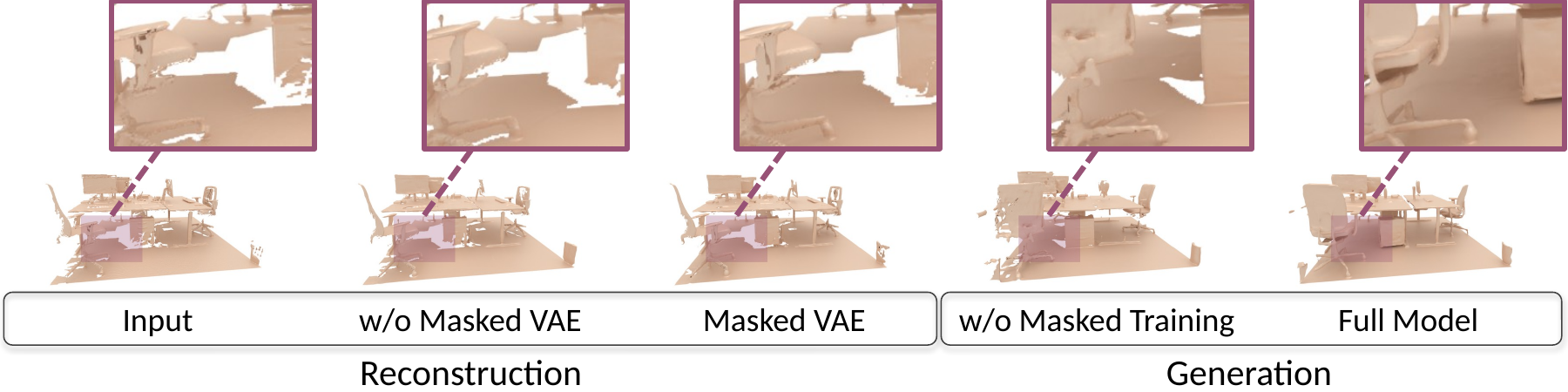}
  \caption{\textbf{Qualitative ablation on masked training.} Comparison of generated scenes with and without masked flow matching during training. Without two stages of masked training, the model tends to generate holes (under the table) following the training incomplete scans.}
  \label{fig:ablation_masked_training}
\vspace{-0.5cm}
\end{figure}

\myparagraph{Impact of Masked Training.}
We ablate the effect of masked training on our final generated scenes in \cref{tab:ablation_masked} and \cref{fig:ablation_masked_training}.
Masked training is crucial to extend generative models to real scans.
With masked training, the sparse VAE better reconstructs input partial 3D scans (\cref{tab:ablation_masked} - Reconstruction), and the SparseDiT synthesizes scene geometry closer to the reference distribution (\cref{tab:ablation_masked} - Generation).
The reason is that without masked training, the model wrongly treats unobserved space (under the table) as empty space and is trained to generate a wrong distribution.

\begin{table}[t]
\centering
\caption{\textbf{Masked training ablation.} Reconstruction metrics (L1, L2, CD) assess the autoencoder and generation metrics (U3D-FPD, VLM) assess the flow matching model; CD is computed on 3D-FRONT only, while others are evaluated across datasets.}
\label{tab:ablation_masked}
\resizebox{\linewidth}{!}{%
\small
\setlength{\tabcolsep}{5pt}
\begin{tabular}{l ccc cc}
\toprule
\multirow{2}{*}{\textbf{Configuration}} & \multicolumn{3}{c}{\textbf{Reconstruction}} & \multicolumn{2}{c}{\textbf{Generation}} \\
\cmidrule(lr){2-4} \cmidrule(lr){5-6}
  & \textbf{L1}{\scriptsize$\times10^{-4}$}$\downarrow$
  & \textbf{L2}{\scriptsize$\times10^{-6}$}$\downarrow$
  & \textbf{CD}{\scriptsize$\times10^{-3}$}$\downarrow$
  & \textbf{U3D-FPD}{\scriptsize$\times10^{2}$}$\downarrow$
  & \textbf{VLM Score}$\uparrow$ \\
\midrule
w/o masked training.   & 3.24 & 7.1 & 10.64 & 20.28 & 5.66 \\
\OURS~(full)          & \textbf{2.01} & \textbf{2.0} & \textbf{9.04} & \textbf{14.56} & \textbf{5.70} \\
\bottomrule
\end{tabular}%
}
\end{table}

\begin{figure}[t]
  \centering
  \includegraphics[width=1.0\linewidth]{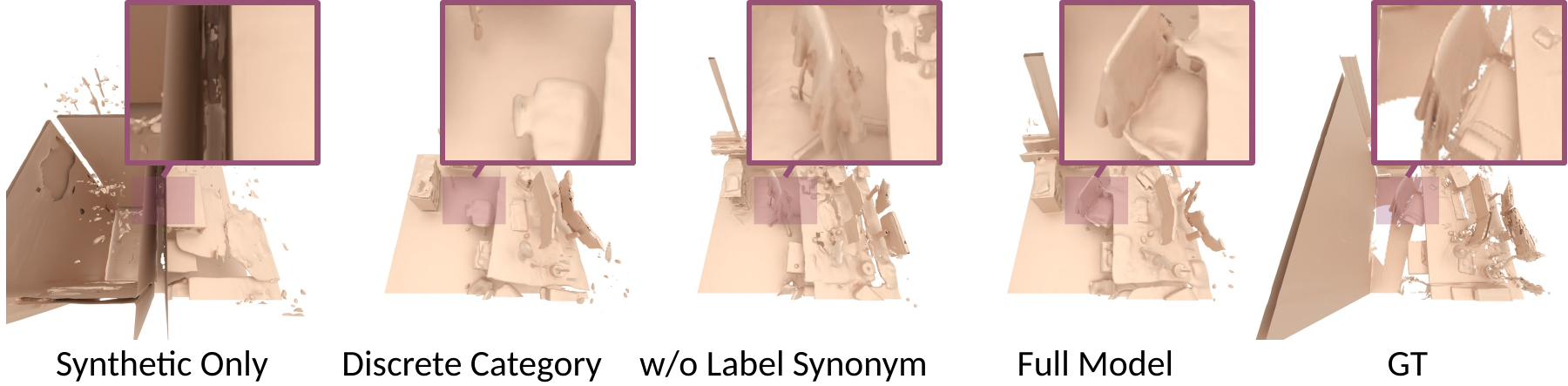}
  \caption{\textbf{Qualitative ablation on real data.} Training the model only on 3D-FRONT~\cite{fu2021_3dfront}, or with limited categories, or without label synonym augmentation, reduces the model's generalization ability to diverse object layouts, like ``clothes on an office chair.''}
  \label{fig:ablation_others}
  \vspace{-0.5cm}
\end{figure}

\myparagraph{Effect of Training on Real Data.}
We ablate the effect of real-world scan data in \cref{tab:ablation} and \cref{fig:ablation_others}. For the synthetic-only setting, we train the model only on the synthetic 3D-FRONT dataset, which has limited object categories and mostly axis-aligned scenes and objects, leading to a large domain gap with the real world. As shown in \cref{fig:ablation_others}, when provided with a complex real layout (\eg, ``clothes on an office chair" as shown in the ground-truth), the model trained with only synthetic data fails to generalize.
The VLM-based score confirms that the full model trained on all datasets exhibit superior overall quality in terms of semantic realism, geometric plausibility, and structural coherence.

\begin{table}[th]
\centering
\caption{\textbf{Ablation study on generation.} We evaluate the contribution of each design choice by removing it from the full model on ScanNet++~\cite{yeshwanthliu2023scannetpp}, ARKitScenes~\cite{arkitscenes}, and 3D-FRONT~\cite{fu2021_3dfront}.
}
\label{tab:ablation}
\resizebox{\linewidth}{!}{%
\small
\setlength{\tabcolsep}{5pt}
\begin{tabular}{l @{\hspace{15pt}} c @{\hspace{10pt}} c @{\hspace{10pt}} c}
\toprule
\textbf{Configuration} & \textbf{DINOv2-FID}{\scriptsize$\times10^{2}$}$\downarrow$ & \textbf{U3D-FPD}{\scriptsize$\times10^{2}$}$\downarrow$ & \textbf{VLM Score}$\uparrow$ \\
\midrule
Synthetic only         & 2.90 & 21.49 & 5.61 \\
Discrete Category      & 3.17 & 21.46 & 5.64 \\
Without Label Synonym      & 2.14 & 18.77 & 5.62 \\
\midrule
\OURS~(full)           & \textbf{1.90} & \textbf{14.56} & \textbf{5.70} \\
\bottomrule
\end{tabular}%
}
\end{table}

\myparagraph{Open Vocabulary Semantic Encodings.}
We ablate the effect of using CLIP embedding~\cite{radford2021clip} instead of one-hot semantic labels which has a fixed number of categories in \cref{tab:ablation} and \cref{fig:ablation_others}. To encode discrete category labels across datasets, we need to map different object labels to the shared object categories. This category mapping is challenging because the same object labels can be categorized into different categories, i.e.,  ``TV'' can also be categorized as ``screen'', ``monitor'', or ``television''. We instead use the CLIP model to embed each object label into a feature vector. This significantly improves the model performance on real datasets. As shown in \cref{fig:ablation_others}, the model using discrete category fails to generate the chair with clothes on it.

\myparagraph{Label Synonym.}
We ablate the effect of using label synonym augmentation in \cref{tab:ablation} and \cref{fig:ablation_others}. To augment the textual object labels, we prompt an LLM~\cite{achiam2023gpt} to provide $20$ variations for each object label in the dataset (prompt details in the supplementary), including: synonyms (``sofa'' → ``couch''), abbreviations (``mobile device'' → ``M''), hyphenated (``bag luggage'' → ``bag-luggage''), plural (``plant'' → ``plants''), and typos (``ceiling'' → ``cielin''). Without the augmentation, the model struggles with diverse object labels as shown in \cref{fig:ablation_others}, where the model only generates a partial office chair.

\section{Conclusion}
\label{sec:conclusion}
We presented \OURS{}, which introduces visibility-aware masked flow matching for 3D scan completion and generation.
Our approach combines a masked sparse VAE that compresses TSDF volumes into compact geometry tokens with a sparse transformer that generates these tokens via flow matching conditioned on semantic layouts.
By training jointly on synthetic and real-world scan data with a masked training strategy that excludes unseen regions from supervision, our model learns to complete missing geometry from incomplete observations.
Experiments on 3D-FRONT, ScanNet++, and ARKitScenes demonstrate that \OURS{} produces coherent, complete scenes and supports controllable generation including text-conditioned 3D scene synthesis in addition to scan completion.

\section*{Acknowledgements}
This work was supported by the ERC Starting Grant SpatialSem (101076253). Matthias Nie{\ss}ner was supported by the ERC Starting Grant Scan2CAD (804724).

\clearpage
\begin{center}\large\bfseries Supplementary Material\end{center}\vspace{1em}
\appendix
\setcounter{figure}{7}
\setcounter{table}{4}
\setcounter{equation}{3}

This supplementary material is organized as follows.
We first provide extended scene completion comparisons (\cref{fig:exp_large_sc_supp}) and more ablation results.
Then, we detail the dataset statistics (\cref{tab:datasets}), scan data processing pipeline, network architecture, and perceptual evaluation protocol.

\section{More Results and Analysis}
\begin{figure}[h!]
  \centering
  \includegraphics[width=1.0\linewidth]{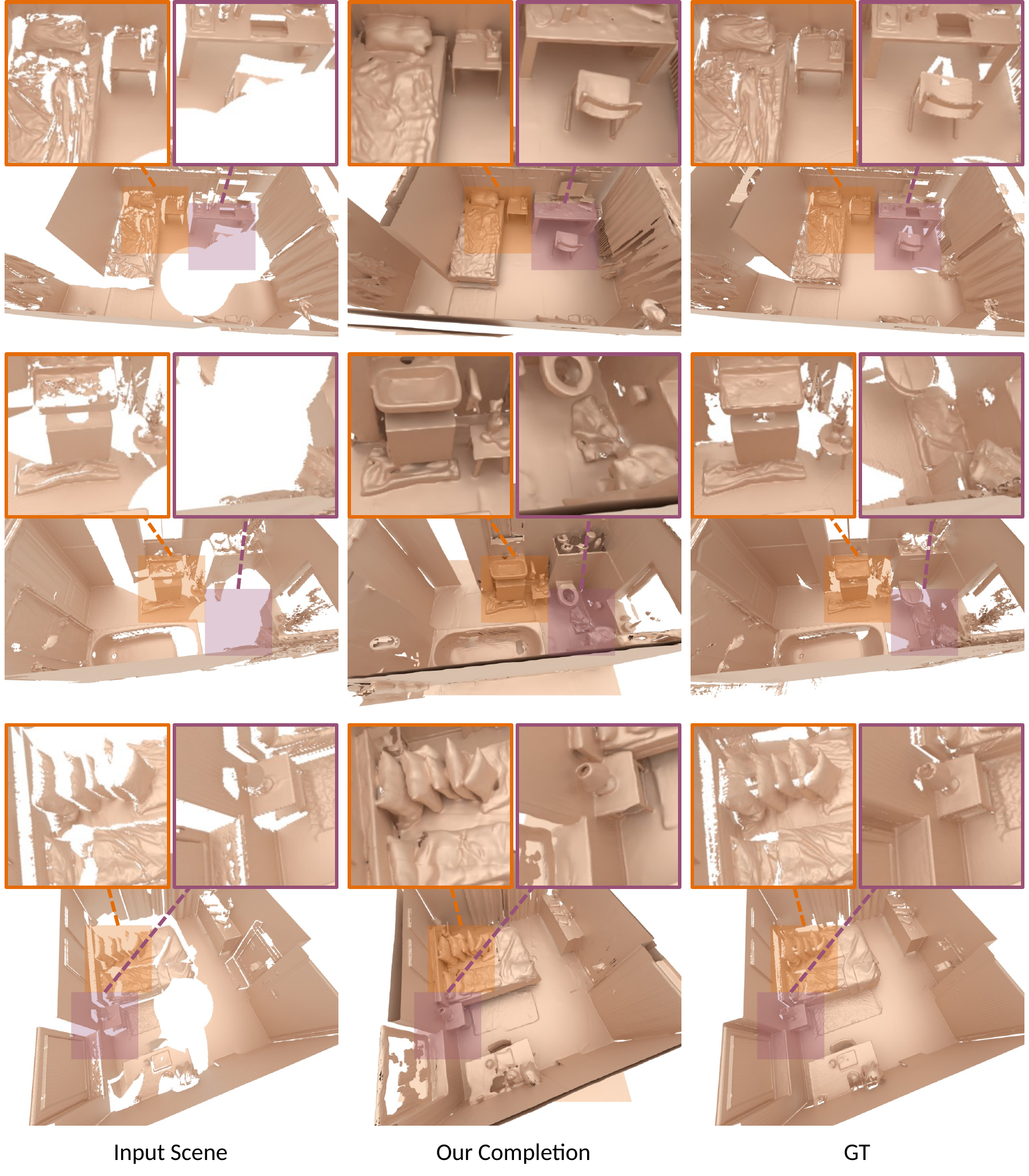}
  \caption{
  \textbf{More 3D scene completion results.} \OURS{} can complete large-scale scenes by generating geometry for unobserved regions across multiple chunks, guided by partial scan conditions injected into the ControlNet branch.
  }
  \label{fig:exp_large_sc_supp}
  \vspace{-2.0cm}
\end{figure}

\myparagraph{More Scene Completion Comparisons.}
We apply the ControlNet-based completion to full house-scale scans from ScanNet++.
Given a partially observed TSDF volume spanning the entire building, we tile it into overlapping patches of resolution $256^3$  and complete each patch independently,
conditioned on the partial chunk observation injected via the ControlNet branch.
Adjacent patches are fused using weighted averaging in overlap regions.
\cref{fig:exp_large_sc_supp} shows more qualitative results: our method successfully infers missing geometry in room-level environments where the partial scan is from single lidar view.

\begin{figure}[t]
  \centering
  \includegraphics[width=1.0\linewidth]{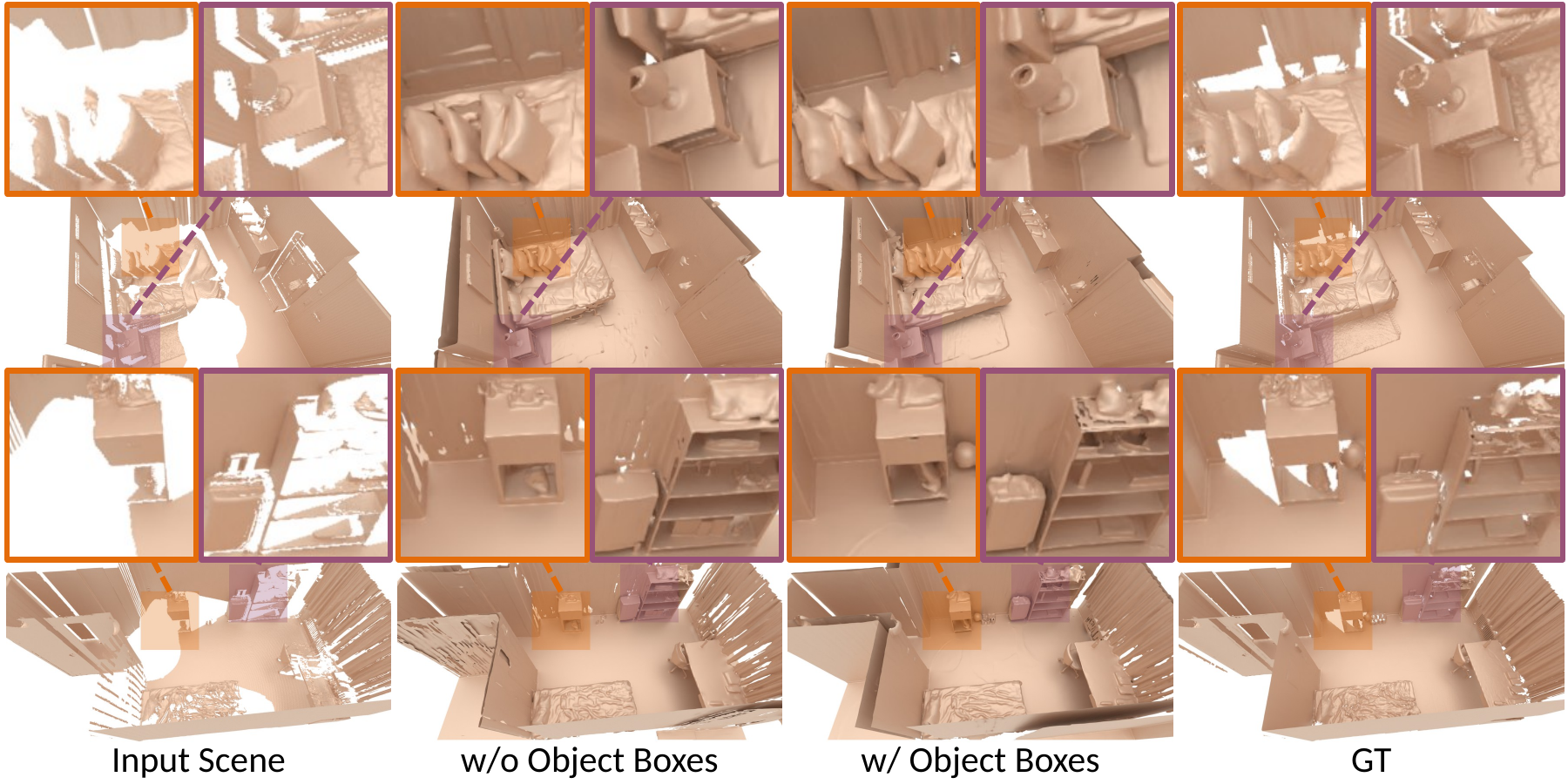}
    \caption{\textbf{Qualitative ablation on object bounding boxes.} \OURS{} achieves comparable completion quality without object bounding box conditioning.}
  \label{fig:ablation_bbox}
\end{figure}

\myparagraph{Qualitative Ablation Studies.}
\cref{fig:ablation_bbox} presents qualitative results for the bounding box conditioning ablation. Explicit object box conditioning is not strictly necessary for 3D scene completion; the model can infer contextual and semantic information directly from the partial scan input to predict missing geometry. Notably, the scenes completed from single-view LiDAR scans appear more geometrically complete than the reference ground-truth scenes, which are reconstructed from a limited number of LiDAR scans.

\section{Dataset Details}
\label{sec:dataset_details}
We train on three indoor scene datasets, summarized in \cref{tab:datasets}.
\textbf{3D-FRONT}~\cite{fu2021_3dfront} provides 6,802 synthetic scenes
with artist-created meshes and dense semantic annotations across 99 categories.
\textbf{ScanNet++}~\cite{yeshwanthliu2023scannetpp} contains 1,006 real-world scenes
captured with high-fidelity LiDAR, offering semantic segmentation over 2,877 categories.
\textbf{ARKitScenes}~\cite{arkitscenes} contributes 928 real scenes
captured with mobile LiDAR, covering 2,466 object categories.
In total, the combined corpus comprises 8,736 scenes with 4,489 unique semantic categories,
spanning both synthetic and real-world indoor environments.

\begin{table}[h]
\centering
\caption{\textbf{Dataset statistics.} Overview of the training datasets used, covering both synthetic (3D-FRONT) and real-world (ScanNet++, ARKitScenes) indoor scenes.}
\label{tab:datasets}
\resizebox{\linewidth}{!}{%
\small
\setlength{\tabcolsep}{5pt}
\begin{tabular}{lc @{\hspace{10pt}} c @{\hspace{10pt}} c @{\hspace{10pt}} c @{\hspace{10pt}} c @{\hspace{10pt}} c}
\toprule
\textbf{Dataset} & \textbf{\# Samples} & \textbf{Synthetic/Real} & \textbf{Object Annot.} & \textbf{Mesh Source} & \textbf{Semantic Seg.} & \textbf{\# Categories} \\
\midrule
3D-FRONT & 6,802 & Synthetic & \checkmark & Artist & \checkmark & 99 \\
ScanNet++ & 1,006 & Real & \checkmark & LiDAR & \checkmark & 2,877 \\
ARKitScenes & 928 & Real & \checkmark & LiDAR & -- & 2,466 \\
\midrule
\textbf{Total} & \textbf{8,736} & -- & -- & -- & -- & \textbf{4,489 (unique)} \\
\bottomrule
\end{tabular}%
}
\end{table}

\begin{figure}[h]
  \centering
  \includegraphics[width=1.0\linewidth]{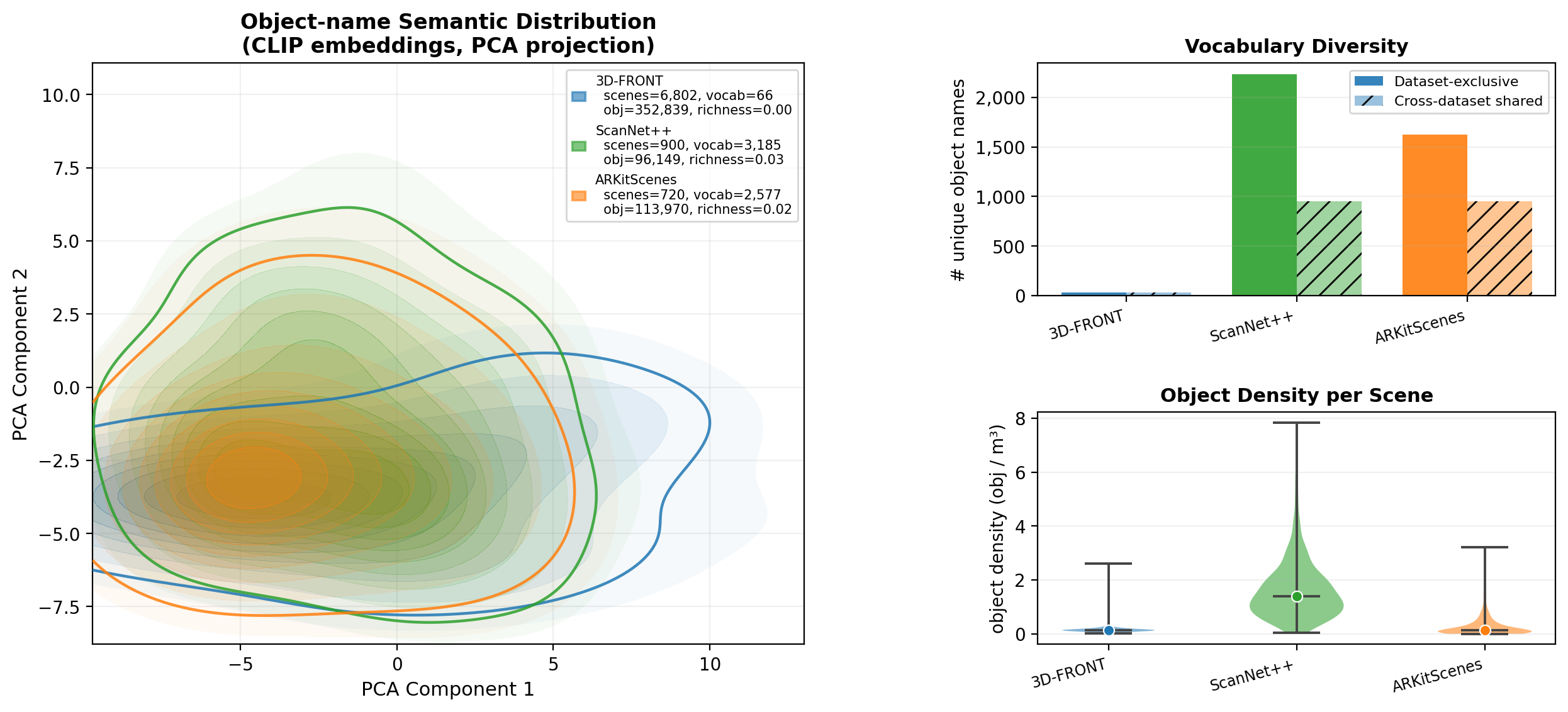}
  \caption{\textbf{Dataset analysis.} Left: object-name semantic distribution (PCA of CLIP embeddings). Top right: vocabulary diversity. Bottom right: object density per scene across 3D-FRONT, ScanNet++, and ARKitScenes.}
  \label{fig:compare_layout_dist}
\end{figure}

\cref{fig:compare_layout_dist}
{visualizes the semantic distribution across the three datasets.
While 3D-FRONT contains only 99 curated furniture categories, ScanNet++ and ARKitScenes span thousands of fine-grained labels covering diverse everyday objects, architectural elements, and cluttered arrangements rarely seen in synthetic data.
Real scans also introduce geometric complexity absent from synthetic scenes: irregular room layouts, partial surface observations from occlusion, sensor noise, and highly variable object densities per room (\cref{fig:compare_layout_dist} - bottom right).
Incorporating this diversity during training is critical for learning robust priors that generalize beyond synthetic environments.
To fully leverage the real scan semantics, we employ LLM-assisted label augmentation (\cref{fig:label_aug_prompt}) to incorporate the long-tailed vocabularies into our semantic schema.}

\section{Scan Data Processing}
\label{sec:data_processing}

All three datasets are converted into a unified TSDF representation via VDB-based volumetric fusion \cite{vizzo2022sensors} with a voxel size of $1.1$\,cm, truncation of $3\times$ the voxel size, and space carving.
For each scene, we extract per-object axis-aligned bounding boxes and semantic names.

\myparagraph{3D-FRONT.}
We use BlenderProc~\cite{Denninger2023} to load scene layouts with 3D-FUTURE~\cite{fu20213d} furniture and export triangulated meshes.
We filter out objects that are empty, contain NaN vertices, have centers far from the scene origin, or exceed plausible furniture dimensions.
Scenes with no remaining structure or furniture after filtering are discarded.
Virtual cameras are sampled per room proportional to floor area, and depth maps are rendered at $512{\times}512$ with $60^{\circ}$ FOV.
The depth images are then fused into TSDF volumes at multiple observation ratios (\eg, $10\%$ and $100\%$) to simulate partial scans.

\myparagraph{ScanNet++.}
We apply the provided rigid alignment transformation to the Faro LiDAR point clouds and directly integrate them into VDB TSDF volumes---no depth rendering is needed.
Object annotations are parsed from the official \texttt{segments\_anno.json}.

\myparagraph{ARKitScenes.}
Mobile LiDAR point clouds are fused into TSDF volumes following the same VDB pipeline.
Object bounding boxes are obtained from CA-1M~\cite{lazarow2024cubify} instance annotations, and floor height is automatically detected from the fused mesh.
We exclude 76 scenes with misaligned LiDAR scans.

\begin{figure}[h!]
    \centering
    \resizebox{0.82\textwidth}{!}{
    \begin{promptbox}{Prompt for Label Augmentation}
        \small
        \textbf{Task:} You are an indoor scene annotation expert. List exactly \texttt{\{n\}} alternative ways ``\texttt{\{label\}}'' might appear as an object category label in datasets like ScanNet++, ARKitScenes, or 3D-FRONT.

        \vspace{2mm}
        \textbf{Important Context:}
        Cover ALL five variation types below. Spread the \texttt{\{n\}} results evenly across them (aim for $\sim$\texttt{\{n\_per\_type\}} per type).

        \vspace{2mm}
        \textbf{Variation Types:}
        \begin{enumerate}[leftmargin=*, noitemsep]
            \item \textbf{Synonyms and Alternative Names}
            \begin{itemize}[label=--, noitemsep]
                \item \eg ``sofa'' $\rightarrow$ couch, settee, divan
                \item ``wardrobe'' $\rightarrow$ closet, armoire, clothes press
                \item ``ceiling lamp'' $\rightarrow$ overhead light, pendant light, ceiling fixture
            \end{itemize}
            \item \textbf{Common Abbreviations Used in Annotations}
            \begin{itemize}[label=--, noitemsep]
                \item \eg ``nightstand'' $\rightarrow$ NS, ``ceiling lamp'' $\rightarrow$ CL, ``television'' $\rightarrow$ TV
                \item ``bag luggage'' $\rightarrow$ bag, ``storage unit'' $\rightarrow$ SU
            \end{itemize}
            \item \textbf{Split / Joined / Hyphenated Spelling Variants}
            \begin{itemize}[label=--, noitemsep]
                \item \eg ``nightstand'' $\rightarrow$ night stand, ``bookshelf'' $\rightarrow$ book shelf, book-shelf
                \item ``tv stand'' $\rightarrow$ tvstand, tv-stand, ``bag luggage'' $\rightarrow$ bag-luggage
            \end{itemize}
            \item \textbf{Singular / Plural / Possessive / Demographic Variants}
            \begin{itemize}[label=--, noitemsep]
                \item \eg ``bins'' $\rightarrow$ bin, ``plants'' $\rightarrow$ plant, ``kids bed'' $\rightarrow$ kid bed
                \item ``childrens desk'' $\rightarrow$ children desk, ``ceiling lamp'' $\rightarrow$ ceiling lamps
            \end{itemize}
            \item \textbf{Frequent Annotation Typos and Misspellings}
            \begin{itemize}[label=--, noitemsep]
                \item \eg ``wardrobe'' $\rightarrow$ wardobe, ``bookshelf'' $\rightarrow$ bookshef, bookshlef
                \item ``nightstand'' $\rightarrow$ nightsatnd, ``ceiling'' $\rightarrow$ cieling, ``cushion'' $\rightarrow$ cusion
            \end{itemize}
        \end{enumerate}

        \vspace{2mm}
        \textbf{Rules:}
        \begin{itemize}[leftmargin=*, noitemsep, topsep=2pt]
            \item One name per line, no numbering, no bullets, no explanations.
            \item All lowercase.
            \item Do NOT repeat ``\texttt{\{label\}}'' itself.
            \item Do NOT include category prefixes like ``furniture:'' or ``object:''.
        \end{itemize}
    \end{promptbox}
    }
    \caption{Prompt template for LLM-based label augmentation. The placeholders \texttt{\{n\}}, \texttt{\{label\}}, and \texttt{\{n\_per\_type\}} are filled at runtime for each category.}
    \label{fig:label_aug_prompt}
\end{figure}

\section{Network Architecture Details}
\begin{figure}[t]
  \centering
  \includegraphics[width=1.0\linewidth]{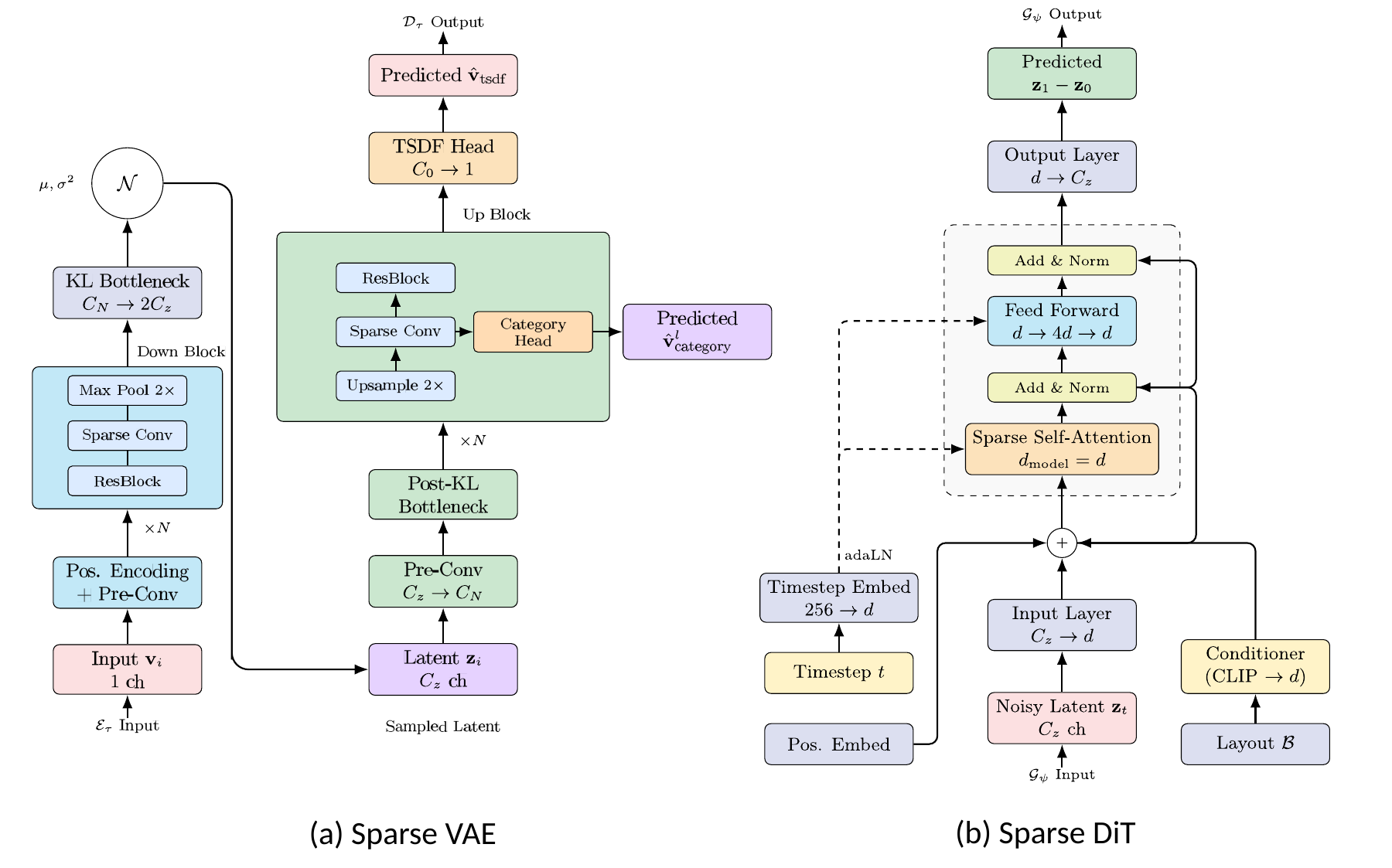}
  \caption{\textbf{Architecture of \OURS.}
  (a)~A masked sparse VAE encodes partial TSDF patches into compact geometry latents via sparse convolutional downsampling, with category and TSDF reconstruction heads in the decoder.
  (b)~A sparse DiT generates geometry latents from Gaussian noise via flow matching, conditioned on CLIP-encoded layout embeddings and timestep information.
  }
  \label{fig:architecture}
\end{figure}

\begin{figure}[t]
  \centering
  \includegraphics[width=1.0\linewidth]{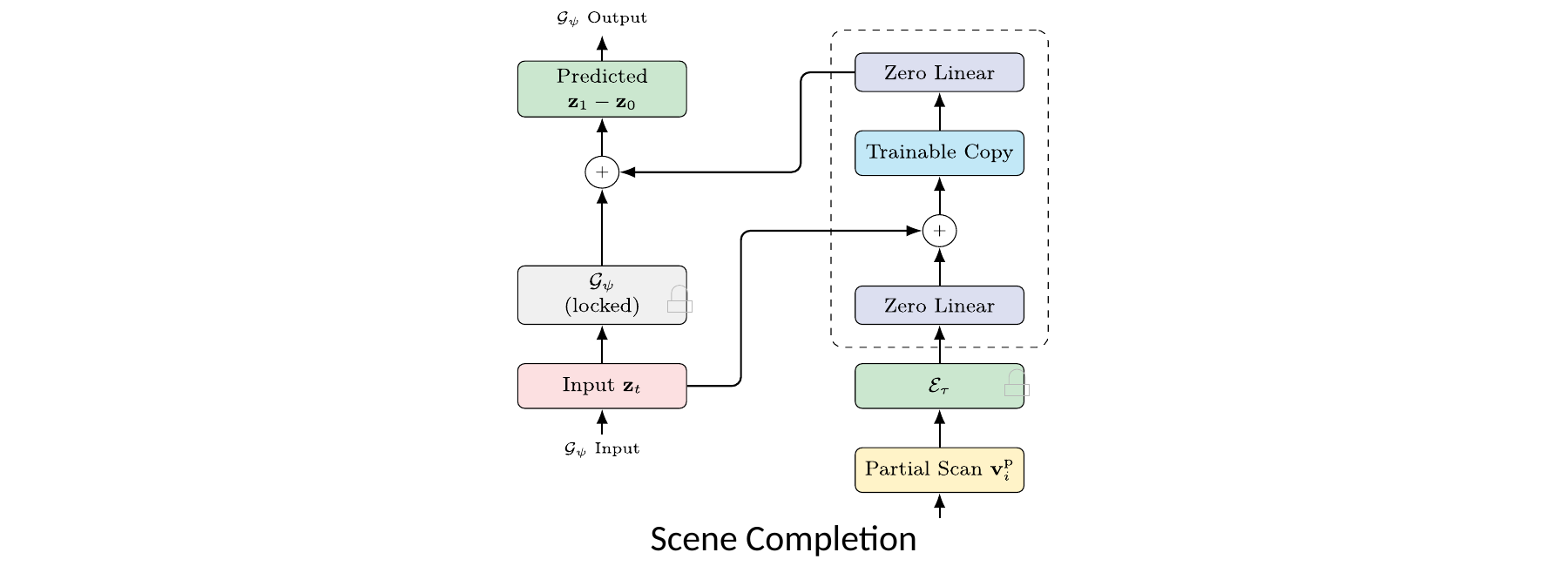}
  \caption{\textbf{Architecture of scene completion.} A ControlNet branch injects partial scan geometry encoded by the frozen VAE encoder into the locked pretrained generator, guiding completion of unobserved regions while the layout condition provides semantic structure.
  }
  \label{fig:architecture_sc}
\end{figure}

\myparagraph{Sparse VAE.}
As shown in \cref{fig:architecture}(a), the sparse VAE compresses a full-resolution TSDF patch into an 8-channel latent with $8{\times}$ spatial downsampling.
The encoder uses three stages of sparse residual blocks (GroupNorm$\to$Conv$\to$SiLU) with $2{\times}$ max-pooling, growing features from 64 to 512, followed by a bottleneck that outputs mean and log-variance for the KL-regularized latent.
Crucially, structure-aware masking retains only known voxels (surface or empty) and replaces unobserved tokens with a learnable empty embedding.
The decoder mirrors this with nearest-neighbor upsampling; at each resolution a structure head prunes empty voxels, and a final geometry head with \texttt{tanh} activation predicts TSDF values.

\myparagraph{Sparse DiT.}
The generator (\cref{fig:architecture}(b)) is a Sparse DiT~\cite{peebles2023scalable, xiang2025structured, wu2025direct3d} with 28 transformer blocks operating on the 8-channel latent.
The input is projected to a 768-dim hidden space with RoPE~\cite{su2024rope} for 3D positional encoding.
Each block contains a 16-head self-attention layer with grouped-query attention (2 KV heads), RMS-normalized QK, and a $4{\times}$-expansion FFN with SiLU.
Timesteps are injected via adaLN modulation; semantic layout conditioning uses CLIP embeddings of object names aggregated per-voxel through instance masks.
We use classifier-free guidance (10\% condition drop), zero-initialized output projection, and EMA weights for inference.

\myparagraph{ControlNet for Scene Completion.}
For scene completion (\cref{fig:architecture_sc}), a control branch---a deep copy of the pretrained Sparse DiT---injects partial scan observations into the frozen generator.
The partial scan latent (from the frozen VAE encoder) is added to $\mathbf{x}_t$ via a zero-initialized projection.
Each of the 28 block outputs passes through a rank-64 zero bottleneck ($768 \to 64 \to 768$) before being added to the corresponding base model output.
All zero projections ensure the model is identical to the pretrained generator at initialization; only the branch parameters are trained.

\section{Limitations}
While \OURS{} demonstrates strong performance on complex, real-world scan geometry, several limitations remain.
Our method inherits the resolution constraints of the underlying TSDF representation and sparse VAE; despite their relatively high resolution, very fine-grained details may be lost during compression.
An interesting future direction is to augment the generated geometry with realistic appearance and material properties, further closing the gap between real-world environments and digital assets.

\section{Perceptual Evaluation Details}

\myparagraph{VLM Scores.}
To systematically quantify perceptual fidelity, we utilize Qwen3-VL-8B-Instruct as an automated evaluator for generated scene chunks. As illustrated in Fig. \ref{fig:vlm_prompt},
the model rates each sample on a 0--10 scale across four key dimensions: Geometric Quality, Structural Logic, Semantic Coherence, and Content Density. This evaluation protocol is specifically designed to distinguish between genuine generative failures and expected artifacts, such as boundary truncation or omitted ceilings, providing a robust, interpretable proxy for judgment of 3D scene plausibility.

\begin{table}[h!]
\centering
\caption{\textbf{VLM-based Perceptual Evaluation.} Comparison of \OURS{} with state-of-the-art baselines on 3D scene generation. Scores are provided by Qwen3-VL-8B-Instruct (0-10 scale), evaluating geometric fidelity, structural logic, semantic coherence, and content density.}
\label{tab:vlm_scores}
\small
\setlength{\tabcolsep}{4pt}
\begin{tabular}{lcccccc}
\toprule
\multirow{2}{*}{\textbf{Method}} & \textbf{Geom.} & \textbf{Struc.} & \textbf{Sem.} & \textbf{Cont.} & \multirow{2}{*}{\textbf{Avg.}} \\
 & \textbf{Qual.} & \textbf{Logic} & \textbf{Cohere.} & \textbf{Dens.} & \\
\midrule
BlockFusion       & 3.82 & 3.92 & 3.43 & 2.85 & 3.51 \\
LT3SD             & 5.22 & 5.48 & 5.14 & 4.97 & 5.20 \\
WorldGrow         & 4.80 & 4.77 & 4.53 & 4.66 & 4.69 \\
\midrule
Seen2Scene (syn.) & 5.27 & 5.79 & 5.74 & 5.62 & 5.61 \\
Seen2Scene (full) & \textbf{5.37} & \textbf{5.84} & \textbf{5.84} & \textbf{5.75} & \textbf{5.70} \\
\bottomrule
\end{tabular}
\end{table}

\myparagraph{VLM-based Evaluation on 3D Scene Generation.} To quantitatively evaluate the perceptual quality and semantic fidelity of the generated 3D scenes, we conduct a comprehensive comparison using the VLM-based scoring metrics described in the main paper. As summarized in \cref{tab:vlm_scores}, Seen2Scene significantly outperforms existing state-of-the-art baselines across all evaluative dimensions. Specifically, our full model, {trained on both synthetic and real scans}, achieves an average score of 5.70, surpassing the previous best-performing method, LT3SD~\cite{meng2025lt3sd}, by a substantial margin. The marked improvements in structural logic (5.84) and semantic coherence (5.84) are particularly noteworthy, as they demonstrate the superior ability of our flow-matching framework to maintain global scene consistency. Furthermore, the performance gain of the ``full'' model over the ``synthetic-only'' variant validates our core hypothesis: training directly on incomplete, real-world scans allows the model to learn robust geometric priors that synthetic datasets alone cannot provide.

\begin{figure}[h!]
    \centering
    \resizebox{0.82\textwidth}{!}{
    \begin{promptbox}{Prompt for Scene Generation Evaluator}
        \small
        \textbf{Task:} You are an expert 3D scene geometry evaluator. You are given a rendered image of a scene chunk (a partial block of a room) derived from a generated scene. Your task is to evaluate the quality of the geometry, specifically focusing on mesh integrity.

        \vspace{2mm}
        \textbf{Important Context:}
        \begin{itemize}[leftmargin=*, noitemsep, topsep=2pt]
            \item These images represent partial views (chunks) of a larger scene.
            \item Ceilings are intentionally omitted; do not penalize for missing ceilings.
            \item \textbf{Boundary Handling:} Do NOT penalize for walls or furniture cut off by the chunk's edge/boundary. This is expected behavior for a partial block.
            \item \textbf{Priority:} Place the highest weight on the integrity of the mesh within the chunk. Objects should not have ``holes,'' missing faces, or ``shredded'' structures unless they are at the very edge of the chunk.
        \end{itemize}

        \vspace{2mm}
        \textbf{1. Assign a score from 0 to 10 for each criterion.} \\
        \textbf{2. Provide a brief justification for your rating.}

        \vspace{2mm}
        \textbf{Rendered Images: [Attached]}

        \vspace{2mm}
        \textbf{Evaluation Criteria:}
        \begin{enumerate}[leftmargin=*, noitemsep]
            \item \textbf{Geometric Quality \& Mesh Integrity:} Are the surfaces solid and complete within the chunk?
            \begin{itemize}[label=--, noitemsep]
                \item Good (8-10): Walls, floors, and objects are watertight and solid. No random holes or ``torn'' surfaces inside the boundary.
                \item Bad (0-3): The mesh looks ``eaten away'' or fragmented. Significant gaps in surfaces.
            \end{itemize}
            \item \textbf{Structural Logic \& Connection:} Do architectural elements connect properly?
            \begin{itemize}[label=--, noitemsep]
                \item Good (8-10): Walls connect logically to the floor. Objects rest on the ground.
                \item Bad (0-3): Walls float in mid-air. Disconnected structural fragments.
            \end{itemize}
            \item \textbf{Semantic Coherence:} Are the objects recognizable and meaningful?
             \begin{itemize}[label=--, noitemsep]
                \item Good (8-10): Objects are clearly identifiable furniture (beds, tables, chairs). They form a logical group.
                \item Bad (0-3): Objects are unrecognizable blobs, messy clusters of geometry, or random debris that doesn't look like furniture.
            \end{itemize}
            \item \textbf{Content Density:} Is the chunk sufficiently populated?
            \begin{itemize}[label=--, noitemsep]
                \item Good (8-10): The chunk contains a reasonable amount of furniture and structure. It feels like part of a room.
                \item Bad (0-3): The chunk is mostly empty void. Contains only a single isolated object (e.g., just one wall and a tiny object) with large empty spaces around it.
            \end{itemize}
        \end{enumerate}

        \vspace{2mm}
        \textbf{Output Format:}
        \begin{tcolorbox}[colback=white, boxrule=0.2pt, arc=1mm, top=2mm, bottom=2mm]
            \ttfamily \scriptsize
            \{ \\
            \hspace*{1em} "scores": \{ \\
            \hspace*{2em} "geometric\_quality": \{"grade": 0, "comment": "..."\}, \\
            \hspace*{2em} "structural\_logic": \{"grade": 0, "comment": "..."\}, \\
            \hspace*{2em} "semantic\_coherence": \{"grade": 0, "comment": "..."\}, \\
            \hspace*{2em} "content\_density": \{"grade": 0, "comment": "..."\}, \\
            \}
        \end{tcolorbox}
    \end{promptbox}
    }
    \caption{Detailed VLM prompts for evaluating perceptual quality of generated 3D scene chunks.}
    \label{fig:vlm_prompt}
\end{figure}

\end{document}